\def\year{2021}\relax
\documentclass[letterpaper]{amsart} 
\usepackage{csl}
\usepackage{times}  
\usepackage{helvet} 
\usepackage{courier}  
\usepackage[hyphens]{url}  
\usepackage{graphicx} 
\usepackage{natbib}  
\usepackage{caption} 
\title[Nonlinear MOR of the QGE through a PI-CAE]{Investigation of Nonlinear Model Order Reduction of the Quasigeostrophic Equations through a Physics-Informed Convolutional Autoencoder}
\author{Rachel Cooper, Andrey A Popov, Adrian Sandu}

\usepackage{amsmath,amsthm,amssymb,amsfonts,subcaption}

\def\!#1{\mathcal{#1}}
\def\*#1{\boldsymbol{\mathbf{#1}}}
\def\|#1{\textnormal{#1}}
\def\##1{\mathfrak{#1}}
\def\x{\mathbf{x}}
\def\z{\mathbf{z}}
\def\fun{\mathbf{f}}

\def\norm#1{\left\lVert#1\right\rVert}

\csltitle{Investigation of Nonlinear Model Order Reduction of the Quasigeostrophic Equations through a Physics-Informed Convolutional Autoencoder}
\cslauthor{Andrey A. Popov, Adrian Sandu}
\cslyear{21}
\cslreportnumber{5}
\cslemail{cyuas@vt.edu, apopov@vt.edu, sandu@cs.vt.edu}

\begin{document}
    
  \csltitlepage   
    
\maketitle
\begin{abstract}
Reduced order modeling (ROM) is a field of techniques that approximates complex physics-based models of real-world processes by inexpensive surrogates that capture important dynamical characteristics with a smaller number of degrees of freedom. Traditional ROM techniques such as proper orthogonal decomposition (POD) focus on linear projections of the dynamics onto a set of spectral features. In this paper we explore the construction of ROM using autoencoders (AE) that perform nonlinear projections of the system dynamics onto a low dimensional manifold learned from data. The approach uses convolutional neural networks (CNN) to learn spatial features as opposed to spectral, and utilize a physics informed (PI) cost function in order to capture temporal features as well. Our investigation using the quasi-geostrophic equations reveals that while the PI cost function helps with spatial reconstruction, spatial features are less powerful than spectral features, and that construction of ROMs through machine learning-based methods requires significant investigation into novel non-standard methodologies.
\end{abstract}

\section{Introduction}
\label{se:introduction}

Machine learning (ML), and particularly neural network (NN) techniques generalize on complex nonlinear manifolds, and extract patterns of information from a given dataset.
However, the recent advent of science-guided \cite{Karpatne2016} machine learning---to which we will refer to as physics-informed (PI)---has revealed the inadequacy of naive machine learning methods for scientific computing applications. 

Similarly, reduced order modeling (ROM) techniques \cite{Noack2011,Mou2021,Quarteroni2015,Benner2015}.  approximate models through the reduction of dimensionality of a given system while both achieving a low approximation error and sufficiently capturing the dynamics of the system with the goal of more efficient computations.
However, while projection-based ROMs have demonstrated success in many applications, depending on the discretization and sampling of observations of the model, the ROM can still fail to capture the dynamics of the underlying system \cite{Noack2008}, leading to the development of closure models, which add some correction to the ROM to achieve better accuracy in these under-resolved situations \cite{Hay2009,Cazemier1998} at the cost of some computational performance and additional resources.

Because of this failure to generalize and capture dynamics in certain systems, and with the increased interest in machine learning (ML), ML-ROM techniques \cite{Daniel2020,San2017,Chen2020PIROM} are now being explored.

The application of PI-ML-ROM schemes branches the fields of ROM and data assimilation, where both theory and observations are used to estimate the model more accurately. This opens ROM into many new applications traditionally covered by data assimilation, \cite{Popov2021a,Popov2021b,Stefuanescu2015,Attia2017}. While data assimilation typically attempts to efficiently combine theory and data within the full space, the introduction of these physics-informed, non-linear ML-ROM schemes can potentially increase performance by allowing for these operations to be conducted within a reduced space instead. Within this paper, we introduce a first step towards the development of a PI ROM framework, the creation of a model reduction scheme that attempts to preserve as much information of the system using a physics-informed loss function in the training process.

Previous work featuring PI loss functions, \cite{Raissi2018_vortex,Muralidhar2018IncorporatingPD,karpatne2018_PIlake}, attempt to match secondary system information attained analytically from the ML model and its outputs. 
We will continue this trend of matching dynamical information from derivatives attained through the training process. 

This work will combine spatial (as opposed to spectral) reduced modeling techniques by utilizing convolution neural-network (CNN) based autoencoders (AE) with PI cost functions that aim to preserve the dynamics of the system we aim to explore: the quasi-geostrophic equations (QGE).

\section{Methodology}
\label{se:methodolgy}
\subsection{Traditional Projection-Based ROM Schemes}
\label{ss:proj_ROM}

Consider a physically accurate, but computationally expensive full order  model (FOM), posed as an initial value problem,
\begin{align}
\label{eq:ODE_model}
   \frac{d}{dt} \,\x(t) = \fun(\x(t)),~~ t \ge t_0, ~~ \x \in \mathbb{R}^{n}.
\end{align}
For example, \eqref{eq:ODE_model} can obtained by the semi-discretization in space of a system of partial differential equations. The state-space of the model is typically very large, $n \gg 1$.

Reduced order models seek to capture the most important dynamical aspects of a full order model \eqref{eq:ODE_model} at a fraction of the computational cost. The traditional approach to constructing ROMs is to project the full order model dynamics \eqref{eq:ODE_model} onto a suitably chosen reduced-order linear subspace \cite{Mou2021}. We discuss the common construction method via Galerkin projections; different Galerkin-ROM approaches differ in their selection of the reduced order subspace \cite{Sastre2008,Quarteroni2015}. In this work we select the reduced order subspace by proper orthogonal decomposition (POD) \cite{Mou2020,San2015}. The process begins by running a full order model simulation \eqref{eq:ODE_model}, and collecting snapshots over $N$ time frames (i.e., saving the full order state at $N$ different time points along the trajectory): 
\begin{multline}
\label{eqn:snapshots}
    \mathbf{X} = 
    \begin{bmatrix}
        \x_1,\x_2, \dots,\x_{N}
    \end{bmatrix} \in \mathbb{R}^{n \times N} \\ \textnormal{where} \quad \x_i = \x(t_{i}), \quad t_{i} > t_{j} ~~\textnormal{for}~~ i>j.
\end{multline}
The snapshot data $\mathbf{X}$ are used to construct the covariance matrix 
\begin{equation}
\mathbf{C} = (N-1)^{-1}\mathbf{X}(\*I_N - N^{-1} \*1_N\*1_N^T)\mathbf{X}^T \in \mathbb{R}^{N \times N},
\end{equation}
where $\mathbf{1}_{N} \in  \mathbb{R}^{N}$ is a column vector of ones. The eigen-decomposition $\mathbf{C}\,\varphi_{i} = \lambda_{i}\,\varphi_{i}$ provides the eigenvalues $\lambda_{1} \geq \lambda_{2} \geq \dots \geq \lambda_{N}$, and the corresponding orthonormal eigenvectors $\{\varphi_{1},\varphi_{2},\dots,\varphi_{N}\}$. The reduced order subspace is chosen to be the space spanned by the $r$ dominant eigenvectors (corresponding to the $r$ largest eigenvalues), $\operatorname{span}\{\varphi_{1},\varphi_{2},\dots,\varphi_{r}\}$, where $r \ll n$. This data-driven construction of basis contrasts with the traditional Galerkin approach, which is universal and not dependent on the problem itself.

Consider the projection matrix with columns given by the basis vectors:
\begin{equation}
\boldsymbol{\varphi} = \begin{bmatrix} \varphi_{1},\varphi_{2},\dots,\varphi_{r} \end{bmatrix}  \in \mathbb{R}^{n \times r}, \quad
\boldsymbol{\varphi}^T\,\boldsymbol{\varphi} = \mathbf{I}_{r \times r},
\end{equation}
with $ \mathbf{I}$ the identity matrix.
The reduced basis is used to approximate solution vectors $\x$ in the full order space by reduced order vector of coefficients $\mathbf{z}$ via the orthogonal projection:
\begin{gather}
\label{eq:gal_ROM}
\mathbb{R}^{n} \ni \x(t) ~~ \Leftrightarrow ~~ \mathbf{z}(t) = \boldsymbol{\varphi}^T\cdot\x(t) \in \mathbb{R}^{r},\\ 
\begin{aligned}
   \x(t) &\approx \tilde{\x}(t) = \boldsymbol{\varphi}\boldsymbol{\varphi}^T\cdot \x(t)\\ 
   \notag
   &= \boldsymbol{\varphi}\cdot \mathbf{z}(t) = \sum_{j=1}^{r} z_{j}(t)\,\varphi_{j}  \in \mathbb{R}^{n},
   \end{aligned}
\end{gather}
where $\tilde{\x}$ is the approximation of $\x$ in the reduced $r$-order subspace. The ROM dynamics that govern the evolution of the  time dependent coefficients $z_{j}(t)$ is obtained by a Galerkin projection of the full order model dynamics \eqref{eq:ODE_model} onto the reduced space:
\begin{equation}
\label{eq:ODE_ROM}
\begin{aligned}
\frac{d}{dt}\, \tilde{\x}(t) =& \boldsymbol{\varphi}^T\cdot\fun(\tilde{\x}(t))
\qquad \Leftrightarrow \\
  \frac{d}{dt} z_j(t) =& \varphi_{j}^T\cdot\fun\left(\textstyle \sum_{j=1}^{r} z_{j}(t)\,\varphi_{j}\right),~~ j = 1,\dots, r.
  \end{aligned}
\end{equation}

\subsection{The Physics-Informed Autoencoder}
\label{ss:PIAE}
ML-ROM basis reduction can be accomplished in several different ways \cite{Daniel2020,Erichson2019}. We will focus on achieving this through an autoencoder framework. Autoencoders are NN-based ML techniques to efficiently learn reduced dimensional approximations of the input data by using both nonlinear reduction (encoding) and reconstruction (decoding) operations~\cite{Kramer1991}. Unlike POD-ROMs, which project spatial data into a spectral space, AE-ROMs directly encode the spatial snapshots within the generated architecture.

Our physics-informed autoencoder approach is based on a philosophy similar to the POD-Galerkin ROM approach. However, instead of constructing a small dimensional linear subspace and using linear projection operators in and out of this subspace \eqref{eq:gal_ROM}, we now employ a nonlinear reduction onto a small dimensional nonlinear manifold, and a nonlinear reconstruction out of this manifold. The nonlinear encoding and reconstruction operators are implemented by an autoencoder.

Much like in the POD-Galerkin ROM (POD-ROM) approach, we start with snapshots \eqref{eqn:snapshots}  of the full order model solutions $\x(t) \in \mathbb{R}^{n}$. Each of the snapshots are reduced onto the small dimensional manifold via the encoder, to obtain reduced size snapshots $\z(t) \in \mathbb{R}^{r}$. Each snapshot is then decoded to obtain a reconstruction $\tilde{\x}(t) \in \mathbb{R}^{n}$ of the original data.

We denote the nonlinear encoding mapping by $\phi(\cdot)$, and the nonlinear decoding mapping by $\psi(\cdot)$:
\begin{align}
\label{eqn:AE-components}
    \begin{split}
        \phi &: \mathbb{R}^{n} \rightarrow \mathbb{R}^{r}, \qquad
        \psi : \mathbb{R}^{r} \rightarrow \mathbb{R}^{n}.
    \end{split}
\end{align}
We assume that the autoencoder component functions \eqref{eqn:AE-components} are (at least) once continuously differentiable. 

For one snapshot $\x(t) \in \mathbb{R}^{n}$ of the full order model state we have:
\begin{equation}
\label{eq:flow}
    \begin{split}
      \z(t)  &= \phi\big(\x(t)\big) \in \mathbb{R}^{r}, \,\,
     \tilde{\x}(t) = \psi\big(\z(t)\big) \in \mathbb{R}^{n} ,
    \end{split}
\end{equation}
where $\z(t)$ is the reduced order representation of $\x(t)$, and $\tilde{\x}(t)\approx \x(t)$ is the full-order space reconstruction. The nonlinear ROM dynamics that govern the evolution of $\z(t)$ is obtained by a projection of the full order model dynamics \eqref{eq:ODE_model} onto the reduced space:
\begin{subequations}\label{eq:AE_Approx}
\begin{equation}\label{eq:AE_Approx-Full}
\frac{d}{dt}\, \z(t) = \phi_{\x}\big(\x(t) \big)\, \frac{d}{dt}\, \x(t) = \phi_{\x}\big(\x(t) \big)\, \fun\big(\x(t)\big),
\end{equation}
where $\phi_{\x}(\cdot)$ is the Jacobian evaluation of the encoder function. This is commonly approximated by substituting the exact state with the reconstruction,
\begin{equation}
\label{eq:AE_Approx-ned}
\frac{d}{dt}\, \z(t) \approx \phi_{\x}\big(\psi(\z(t))\big) \, \fun\big(\psi(\z(t))\big).
\end{equation}
\end{subequations}

A traditional autoencoder learns the nonlinear mappings to the reduced order space and back into the full space using only full-order model state-space data. Without any knowledge of the underlying physics of the governing dynamics \eqref{eq:ODE_model}, training may lead to reconstructions that break physical consistencies, or may display local accuracy but have poor generalization performance when the ROM is used to propagate a reduced order solution through time. 

We propose imposing the physical information of the governing equations  \eqref{eq:ODE_model} via a physics informed loss (PI-loss) function. For each snapshot $\x(t) \in \mathbf{X}$ we seek to minimize both the state reconstruction by the ROM autoencoder:
\begin{align}\label{eq:State_minimize}
    \min_{\phi,\psi} \norm{\tilde{\x}(t) - \x(t)}_{2}^{2} = \min_{\phi,\psi} \norm{\psi(\phi(\x(t))) - \x(t)}_{2}^{2},
\end{align}
as well as the error incurred by the reduced order approximation of the governing dynamics \eqref{eq:AE_Approx}. We take the (practical) approach to penalize the mismatch between the full order and the reduced order dynamics in the full order space, specifically:
\begin{subequations}
\label{eq:RHS_penalty}
\begin{align}\label{eq:RHS_minimize}
    \min_{\phi,\psi} \norm{\frac{d}{dt}\tilde{\x}(t) - \frac{d}{dt}\x(t)}_{2}^{2}
\end{align}
where $\norm{\cdot}_{2}^{2}$ is the squared $L^2$ norm. The time derivative of the reconstruction $\tilde{\x}$ is represented using the components of our autoencoder as follows:
\begin{equation}\label{eq:RHS_derivation}
    \begin{split}
	\frac{d}{dt}\tilde{\x}(t)& = \frac{d}{dt}\psi\big(\phi(\x(t))\big)\\
    & = \psi_{\z}\big(\phi(\x(t))\big)\,\phi_{\x}\big(\x(t)\big)\,\fun\big(\x(t)\big)\\
    & = \psi_{\z}\big(\z(t)\big)\,\phi_{\x}\big(\x(t)\big)\,\fun\big(\x(t)\big),
    \end{split}
\end{equation}
\end{subequations}
where $\psi_{\z}\left(\cdot\right)$ is the Jacobian of the decoder function, and $\phi_{\x}(\cdot)$  the Jacobian  of the encoder function, as discussed above. The computation of the right-hand side in \eqref{eq:RHS_derivation} involves matrix-vector products between the Jacobians of the nonlinear functions that compose the autoencoder and the right-hand side of the input. 

In this work we opt to use the true snapshots and dynamics within the minimization \eqref{eq:RHS_minimize}, rather than the approximation \eqref{eq:AE_Approx-ned} based on reconstruction:
\begin{equation*}
\min_{\phi,\psi} \norm{\psi_{\z}\big(\z(t)\big)\,\phi_{\x}\big(\tilde{\x}(t) \big)\,\fun\big(\tilde{\x}(t) \big)- \fun\big(\tilde{\x}(t) \big)}_{2}^2,
\end{equation*}
for two reasons. First, the computational cost of training would considerably increase by implementing the approximate right hand side evaluations. The proposed minimization of the dynamics error \eqref{eq:RHS_minimize} only requires the full order right hand side to be evaluated once, before training, at the full order solution snapshots. When reconstructed snapshots $\tilde{\x}(t)$ are used, then at every step of the training process the right-hand side of the full order dynamics will have to be recomputed. Second, we wish to minimize the error in the projected dynamics; by constructing the penalty based on the true dynamics rather than the reconstructed one we guarantee that the learning model is correctly selecting relevant features present in the true data, rather than inferring these features from possibly incorrect reconstructions.

While obtaining the full Jacobians $\psi_{\z}\left(\cdot\right)$ and $\phi_{\x}(\cdot)$ directly is a computationally costly task, modern automatic differentiation packages included in popular NN frameworks such as Pytorch \cite{PyTorch2019} can compute vector-Jacobian and Jacobian-vector products  of these complex non-linear mappings without the need to build the full Jacobian through automatic differentiation. This efficient process makes it feasible to use these physics informed metrics \eqref{eq:RHS_penalty} in the loss function, while keeping the computational costs reasonable within the training process. 

With our minimization criteria expressed in terms of model computations, we arrive at the proposed PI-loss function used to train our AE-ROM:
\begin{equation}\label{eq:PI_Loss}
\begin{split}
	    \mathcal{L}
        = \sum_{i=1}^{N}& \Big(  \norm{\tilde{\x}_{i} - \x_{i}}_{2}^{2} + \\ 
        &\quad \lambda\norm{\psi_{\z}\left(\z_{i}\right)\,\phi_{\x}(\x_{i})\,\fun\left(\x_{i}\right) - \fun(\x_i)}_{2}^{2} \Big).
\end{split}
\end{equation}
The regularization constant $\lambda$ controls the importance placed on the minimization of the physical dynamics of the reconstruction. Within this work, we use a regularization of $\lambda = 10^{-4}$, making the matching of the dynamics a weak physical constraint in the training process as to not fully disrupt the natural training process of the AE-ROM. The choice of constant $\lambda$ is strongly motivated by the mismatch between the distribution of the data and the distribution of the dynamics of the data. Related work \cite{Azencot2020,Lee2019} makes similar use of PI-loss AE-ROMs. These PI-losses seek to guide the selection of non-linear basis transformations to maintain the relevant physical quantities when propagated though the reduced order model \cite{Azencot2020}, or through an iterative method such as Newton’s method \cite{Lee2019}. 

\section{Experimental Setup}
\label{se:exp_setup}

\subsection{The physical model}
\label{ss:QGE_explain}
We evaluate the performance of our proposed Physics-Informed Autoencoder ROM (PI-AE-ROM) model on the single-layer two-dimensional quasi-geostrophic equations (QGE) \cite{San2015}. The QGE describe the motion of rotating flows seen in the interactions between oceanic and atmospheric circulations and is a frequent test problem within fluid flow ROM \cite{Mou2020,San2018}. The streamfunction representation of the QGE is:
\begin{align}\label{eq:QGE}
    \begin{split}
        \frac{\delta \omega}{\delta t} + J(\omega,\varphi) - Ro^{-1}\frac{\delta\varphi}{\delta x} &= Re^{-1}\Delta \omega + Ro^{-1}F, \\
        \omega(t,x,y) &= -\Delta\varphi(t,x,y),
    \end{split}
\end{align}
where $\varphi$ and $\omega$ are the streamfunction and vorticity of the system defined over time in the x and y directions, respectively, $Re$ is the Reynolds number, $Ro$ the Rossby number, and $F$ an external forcing to the system. 
We use $Re=450$, $Ro=0.0036$, and the symmetric double gyre forcing $F = \sin(\pi(y-1))$ \cite{Mou2020,San2015}.
A time discretization of one day representing a change in model time of $\Delta t = 0.0109$ is used. The initial condition used within the evaluations is a representative model state; generated by running a model with zero initial conditions for roughly 1,000 days---time $t=10$, when the solution becomes stable.

We further discretize QGE over a spatial grid resolution of 127 equidistant samplings in the $x$-direction, and 63 samples within the $y$-direction via a second order finite difference scheme, where the non-linear term is approximated by the Arakawa approximation. Thus, each snapshot in time is a field $\x(t) \in \mathbb{R}^{127 \times 63}$, resulting in a total full-space data size of $n=8001$. This high order full space makes this QGE discretization a prime candidate for ROM applications.

The snapshot dataset is generated from trajectories of the full order system. From the initial condition of the system, a primary training dataset is made on a timescale of six months, with a total of $N=2001$ samples of the QGE system representing the timeframe. The models are all trained and validated on this dataset, and analysis will be conducted to judge the validity of the models in reducing physical inconsistencies in a generalized sense. 

We compare the PI-AE-ROM against two reference reduced order models. To assess performance against traditional ROM schemes, a classical POD-ROM model is constructed via the methodology discussed in section \ref{ss:proj_ROM}. To assess the benefits of introducing physical information in the loss function \eqref{eq:RHS_minimize} we compare the PI-AE-ROM scheme against an AE-ROM of similar construction, but where the loss function is based solely on the state-space data \eqref{eq:State_minimize}. We summarize the construction of these different ROMs below.

\subsection{Classical POD-ROM model}
\label{ss:POD_ROM_setup}
The construction of the POD-ROM model used as a reference within this work is obtained through the ODE Test Problems \cite{roberts2021ode,otpsoft} package. The basis vectors used to project the full space data into the reduced order space are obtained through the scheme discussed in section \ref{ss:proj_ROM}. Furthermore, a reduced order representation of the QGE dynamics is created through a Galerkin projection of the state variables $\omega$ and $\varphi$ onto the reduced space as well via \eqref{eq:gal_ROM}. The quadratic reduced order system is defined as:
\begin{equation}\label{eq:G-ROM_Approx}
    \frac{d}{dt} {\mathbf{z}} = \mathbf{b} + \mathbf{A}\,\mathbf{z} + \mathbf{z}^{T}\,\mathbf{B}\,\mathbf{z},
\end{equation}
where $\mathbf{b}$, $\mathbf{A}$, and $\mathbf{B}$ represent the constant, linear, and quadratic terms, respectively  \cite{Wang2012}. The reduced dimensional order $r$ is selected via the number of dominant eigenvectors $\varphi$ chosen in the construction and projection process within section \ref{ss:proj_ROM}.

\subsection{AE-ROM models}
\label{ss:PIAE_setup}
\begin{figure*}[t]
    \centering
    \includegraphics[width=0.9\linewidth]{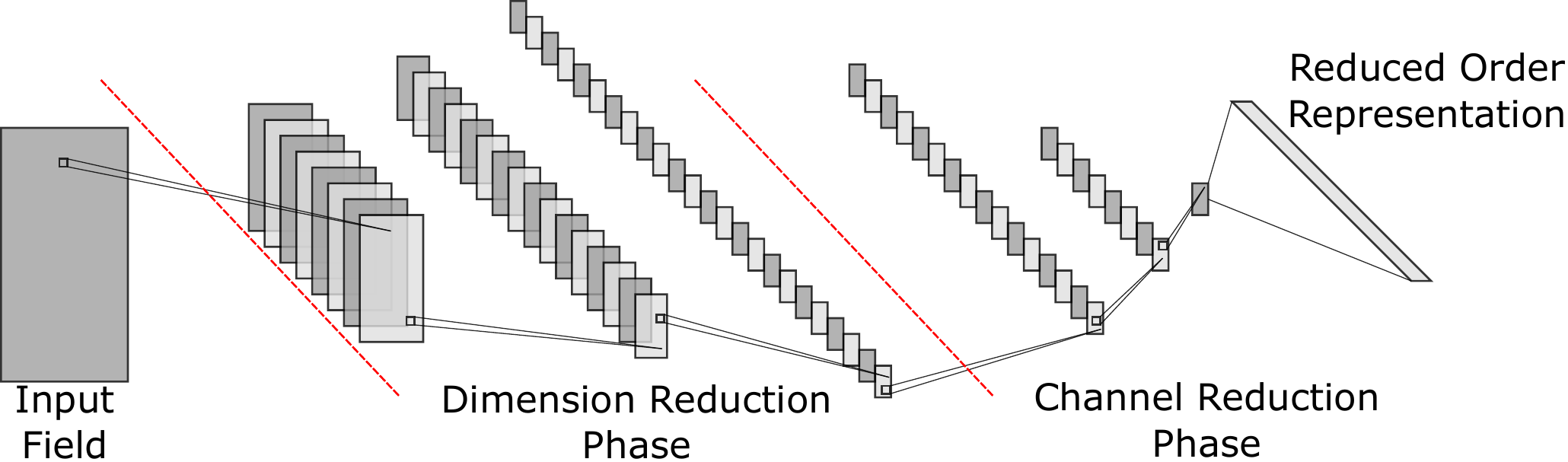}
    \caption{Illustration of the encoder section of the AE framework. From left to right: the input within our full space, the dimension reduction phase where the size of an individual channel is reduced and the total number of channels increased, the channel reduction phase where now the dimensionality of channels is fixed and the total number of channels are reduced, and the encoded information which is attained through a linear transformation of the final output of the channel reduction phase.}
    \label{fig:Encoder_Breakdown}
\end{figure*}

The QGE state space \eqref{eq:QGE} consists of 2D fields where entries are locally related to each other; for this reason, a CNN structure is used as the AE construction framework. The AE is built within four mirrored phases: a dimensional-shrinking phase, a channel-shrinking phase, then (in the mirror) a channel-growth phase and finally a dimensional-growth phase. Within dimensional phases, the number of channels changes in an inverse relationship to the dimensionality, so that if the order is being reduced, the number of channels are being expanded to compensate. For the channel phases, the order of a single channel is kept constant, while the number of channels is expanded or reduced. This allows for a rough reduction of the model, with the specific order of the reduced order space being determined with a linear layer transformation between the channel phases. The encoder portion of this framework is illustrated in figure~\ref{fig:Encoder_Breakdown}.

The purpose of this scheme is to adequately capture the characteristics of the full-order data.  
Because of the nature of our proposed loss function and the QGE system dynamics themselves, trying to capture most of the dimensionality reduction within the 2D structure of convolutional layers is ideal to further enhance the found relationships within the functions right hand side we wish to capture. 
With the majority of the model reduction captured within the convolutional space, we fine-tune the reduced order via a simple linear mapping with an affine layer of controllable size. The size of this layer represents the reduced order representation of the model, e.g., if the layer is specified to have 10 neurons, the reduced order representation with be of order $r=10$.

The construction of the nonlinear operators that project the QGE snapshots into the reduced order space and back into the full space needs to be complemented by the reduced order dynamical representation of the QGE. As the construction of an optimal reduced order dynamical representation is not the focus of this work, in the numerical experiments we use the reduced order dynamics given by \eqref{eq:AE_Approx}.

\subsection{Training with two different loss functions}
We consider two AE-based ROMs. The first one, named PI-AE-ROM, is trained using  PI-guided loss function \eqref{eq:PI_Loss}.
The second one, named the MSE-AE-ROM, has the same architecture but does not include the physics term in the loss function to constrain the AE. The second model is trained using a classical data-only mean squared error (MSE) loss:
\begin{equation}\label{eq:MSE_AE}
    \text{MSE}(\mathbf{\tilde{X}}) = \frac{1}{N}\sum_{i=1}^{N}\left(\x_{i} - \tilde{\x}_{i}\right)^{2}.
\end{equation} 
Both AE models are built within the PyTorch framework \cite{PyTorch2019}, using the ADAM optimizer \cite{Kingma2017}, and ELU activations \cite{Clevert2016} used between convolutional layers. To compare the training performance of both AE models, both the PI-loss \eqref{eq:PI_Loss}, and the MSE loss metrics will be reported at every tenth epoch.

\subsection{Evaluation Techniques}
\label{ss:eval_techniques}
All three models will be tested over a range of reduced dimensional orders, selected to be $r=\{5, 10, 25, 50, 100\}$. These dimensional orders will be used to better understand the models' validity over a range of different feature compressions.

To compare the training performance of the two AE models, both loss metrics will be reported to gauge both the validity of the PI-loss metric towards guiding the training, and to compare the speed of convergence over the training epochs.

To measure the reconstruction error between all three models, we use the error metric detailed in \cite{Mou2020},
\begin{equation}\label{eq:energy_vel}
    E_{v}(\tilde{\x}) = \left(\frac{\partial}{\partial x} \tilde{\x}\right)^{2} + \left(\frac{\partial}{\partial y} \tilde{\x}\right)^{2}
\end{equation}
where $\frac{\partial}{\partial x}$ and $\frac{\partial}{\partial y}$ are spatial derivatives calculated by second order spatial finite difference s, representing measures within the directional velocity in the $x$ and $y$ directions of the QGE.

This metric measures both the general reconstruction of the spatial data, but also validates that the constructed ROMs have selected relevant features towards the propagation of the model. This error metric is then compared to the reference energy via a ratio.

Finally, the stability of the projection schemes will be compared through propagating the resulting ROMs with the RK4 Runge-Kutta scheme~\cite{Butcher1996_RK}. At every evaluation step of the RK4 propagation, the full-space snapshots are projected into the reduced space via the relevant POD-ROM or AE-ROM model. The Runge-Kutta step is then taken in the reduced space using the ROM QGE approximations, \eqref{eq:G-ROM_Approx} and \eqref{eq:AE_Approx}, the result is then projected back into the full space. In total, 50 steps are taken with the RK4 scheme to propagate the snapshot datasets half a physical day forward in time. These datasets are denoted as $\mathbf{X}'$ and $\tilde{\mathbf{X}}'$ for the final RK4 solutions of the true dataset, and ROM model approximations respectfully.

Within both the MSE-AE-ROM and PI-AE-ROM models, the direct construction of a reduced order dynamical representation is not achieved within this work. Instead \eqref{eq:AE_Approx} is used to approximate the dynamics within the reduced order representation. Within the POD-ROM, \eqref{eq:G-ROM_Approx} is used as the projected dynamics within the reduced space.

The reference trajectory $\mathbf{X}'$ is generated in the full space using the true dynamics \eqref{eq:QGE}. The ROM trajectories are compared via a relative $L^{2}$ error metric,
\begin{equation}\label{eq:metricloss}
	\ell(\mathbf{X}', \tilde{\mathbf{X}}') = \sum_{i=1}^{N} \frac{\norm{\x_{i}' - \tilde{\x}_{i}'}_{2}^{2}}{\norm{\x_{i}'}_{2}^{2}},
\end{equation}
to describe the difference between the reference ``true'' solution and the reconstruction. 

\section{Results}
\label{se:results_AE}

\subsection{AE-ROM Training Results}
\label{ss:AE_ROM_Training}

\begin{table}[t]
    \centering
    \caption{Comparison of PI-loss metrics \eqref{eq:PI_Loss} of both AE-ROM models, at final training epoch 50. (Lower is better)}  
    \begin{tabular}{|c|cc|}
        \hline
        r & MSE-AE-ROM & PI-AE-ROM \\
        \hline
        5 & 2.702970 & \textbf{1.869922} \\
        10 & 1.489654 & \textbf{0.927310} \\
        25 & 1.249153 & \textbf{0.595015} \\
        50 & 0.733397 & \textbf{0.467942} \\
        100 & 0.579573 & \textbf{0.352489} \\
        \hline
    \end{tabular}
    \label{tab:Training_PI}
\end{table}

\begin{table}[t]
    \centering
    \caption{Comparison of MSE metric losses \eqref{eq:MSE_AE} of both AE-ROM models, at final training epoch 50. (Lower is better)}
    \begin{tabular}{|c|cc|}
        \hline
        r & MSE-AE-ROM & PI-AE-ROM \\
        \hline
        5 & \textbf{0.729787} & 0.798952 \\
        10 & 0.235952 & \textbf{0.228055} \\
        25 & 0.191920 & \textbf{0.125057} \\
        50 & 0.098768 & \textbf{0.096549} \\
        100 & 0.069425 & \textbf{0.060848} \\
        \hline
    \end{tabular}
    \label{tab:Training_MSE}
\end{table}

Each AE-ROM model is given 50 epochs to converge, with the final training losses reported in tables \ref{tab:Training_PI} and \ref{tab:Training_MSE}. For the PI-Loss metric, \eqref{eq:PI_Loss}, the PI-AE-ROM achieves roughly half of the error as the MSE-AE-ROM, indicating that this is new information given to the model and not something that can be implicitly inferred in the learning process. Within the MSE loss, the PI-AE-ROM again achieves a lower error in all but the most restrictive of the ROM reductions, the fifth order encoding ($r=5$).

\subsection{Energy Preservation}
\label{ss:Energy_Preservation}
We now move towards comparisons between all three models, the POD-ROM, MSE-AE-ROM, and PI-AE-ROM. Beginning with the reconstruction accuracy, we measure the performance via a ratio of \eqref{eq:energy_vel}. As this is a ratio on how much dynamical information is preserved in the ROM reconstruction compared to the true dynamical information; results closer to $1$ represent more accurate models, with lower values indicating information loss. The results are summarized in table \ref{tab:Energy}.

\begin{table}[!htb]
    \centering
    \caption{Energy ratio (reconstruction vs true) for selected reduced dimensional orders. Values near one are better.}
\begin{tabular}{|c|ccc|}
    \hline
    r & POD-ROM &MSE-AE-ROM & PI-AE-ROM \\
    \hline
    5 & 0.779468 & \textbf{0.898383} & 0.880781\\
    10 & 0.904329 & \textbf{0.972391} & 0.958915\\
    25 & \textbf{0.967073} & 0.987748 & 0.922606\\
    50 & \textbf{0.986692} & 0.955068 &  0.909075\\
    100 & \textbf{0.996369} & 0.985244 & 0.965877 \\
    \hline
\end{tabular}
    \label{tab:Energy}
\end{table}

Both AE-ROMs perform better than the traditional POD-ROM within the smaller reduced order spaces of total dimensional order five, and ten. Within the less restrictive reduced dimensional compositions though, we instead see the POD-ROM claim the more accurate reconstructions. This result shows the applicability for non-linear ROM schemes such as the AE-ROMs introduced here. When a small ($r \ll n$) reduced order space is desired, the construction of a more computational expensive nonlinear scheme may be considered over a traditional projection-based ROM.

One significant observation from these energy conservation results is the performance on the metric between the two AE-ROM models. Although the PI-AE-ROM performs better in the training metrics, we see that for all reduced dimensional orders, it loses more dynamical energy information compared to the MSE-AE-ROM.

\subsection{Reconstruction Accuracy}
\label{ss:neconstruction_Accuracy}

\begin{figure}[t]
    \centering
    \includegraphics[width=0.99\linewidth]{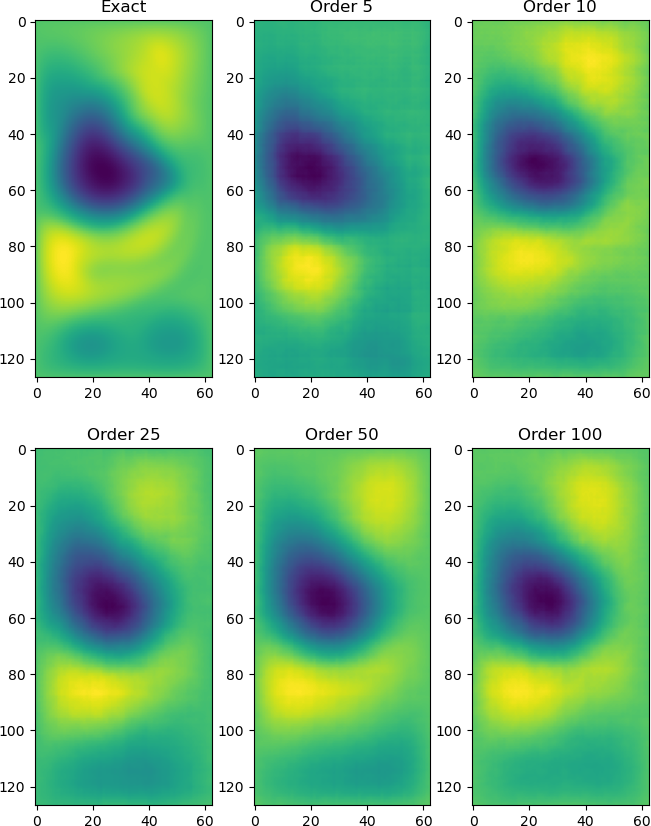}
    \caption{Comparisons of reconstructions between reduced dimensional orders of the PI-AE-ROM, against the ``true'' dynamics (upper left).}
    \label{fig:PI_Comp}
\end{figure}

We next qualitatively compare the reconstructions achieved by PI-AE-ROM models are shown within figure \ref{fig:PI_Comp}. The figure begins with the true dynamics of the QGE system (section \ref{ss:QGE_explain}), then progresses from the smallest reduced dimensional order, $r=5$ to the largest order of $r=100$.

For the larger reduced dimensional order models ($r=50$ and $r=100$), we achieve comparable qualitative reconstructions, where POD-ROM model, being a spectral method would show smooth reconstruction. Within the smaller orders, while the results in table \ref{tab:Energy} indicate that we capture and reconstruct more of the model features within the AE-ROM schemes, we see a tiling within the visualized fields, with sharp discontinuities between local results. 

These qualitative results clearly indicate that using only local features and reconstructions would lead to non-physical approximations of the QGE states.

\subsection{ROM Propagation}
\label{ss:RK_4_results}

Finally, we examine the results of the ROM propagation scheme detailed in section \ref{ss:eval_techniques}. We apply the RK4 method to all snapshots within both the present and future reference data sets constructed. Table \ref{tab:RKNear} reports the results for the training data set.

\begin{table}[!htb]
    \centering
    \caption{Relative solution error \eqref{eq:metricloss} for current time interval RK4 propagation experiment.}
\begin{tabular}{|c|ccc|}
    \hline
    r & POD-ROM &MSE-AE-ROM & PI-AE-ROM \\
    \hline
    5 & \textbf{0.472872}   & 2.427977 & 1.192203  \\
    10 & \textbf{0.293625}  & 17.258518 & 25.309463   \\
    25 & \textbf{0.203355}  & 24.729580 & 3.191429  \\
    50 & \textbf{0.189302}  & 9.895016 & 1.032118  \\
    100 & \textbf{0.054356} & 1.930263 & 2.599640  \\
    \hline
\end{tabular}
    \label{tab:RKNear}
\end{table}

We see that our simple dynamical approximation of the AE-ROM reduced order dynamics, \eqref{eq:AE_Approx}, performs much worse than the POD approximation, never achieving a relative $L^{2}$ error under a value of one for any reduced dimensional mode. 

Interestingly, we see high spikes of error within some orders, such as the results in both the current and future time intervals for the MSE-AE-ROM with a reduced dimensional order of $r=25$. We see these highly anomalous results paired in the two time intervals, indicating it is an issue with the AE-ROM model, rather than with the RK4 scheme employed, or the specific dataset used. While deep analysis was not performed as the efficient construction of a right-hand side approximation was not the focus of this paper, one can think that this may occur from a poorly formed encoder, $\phi(\cdot)$, leading to a poor Jacobian for the QGE ROM approximation used \eqref{eq:AE_Approx-ned}.

\section{Discussion}
\label{se:disussion}
This work investigates a physics-informed autoencoder approach for efficient nonlinear model order reduction. The proposed PI-AE-ROM model uses both observed model trajectory data and dynamical  (right hand side) information within its loss evaluation metric,  to enable the reconstruction process to select features and relationships that preserve the dynamics of the system. We compare this model to both an autoencoder of similar architecture but with a classical data-based-only loss function, and a POD-ROM solution on the single layer quasi-geostrophic equations.

Using the velocity energy metric \eqref{eq:energy_vel}, we find that the nonlinear AE-ROM models outperform the POD-ROM for smaller orders but fail to capture as much energy as the POD-ROM for larger orders. Comparing the two AE-ROMs to each other, an interesting trend emerges. While the proposed PI-AE-ROM achieves better training scores on both the PI-loss and MSE metrics, we see that in the evaluations of the reconstructions, the performance of the PI-AE-ROM is always lower than the MSE-AE-ROM.

This somewhat unexpected result, where the AE-ROM model that performs better on the training metrics performs worse on the reconstruction evaluations, indicates that the MSE and PI-loss metrics are insufficient to reduce dynamical inconsistencies experienced within the ROM reconstructions of the QGE system. Other physics-informed application results \cite{Yang2020_enforcing} indicate that this may be a common issue. While a training metric may improve convergence and accuracy on a specific physical measurement, poor results can still occur with respect to the conservation of other physical properties.


This work uses a single regularization parameter value ($\lambda = 10^{-4}$) of  physics informed information within the loss criteria,\eqref{eq:PI_Loss}. As the additional information of this PI metric does appear to speed up the training process, future work could also focus on selecting $\lambda$ to attain an optimal speed-up of the training process.




Additionally, our experiments with propagating (running forward in time) the AE-ROM dynamical approximations \eqref{eq:AE_Approx-Full} show poor results when compared to the POD-ROM model, with no AE-ROM construction achieving a relative $L^{2}$ error of less than $1$. Additionally, the AE-ROM dynamics generates anomalous spikes for several configurations, such as the MSE-AE-ROM constructed with a reduced order dimensionality of 25 reporting an average $L^{2}$ score of $24.513$ from the two tested time intervals. As mentioned above, this could be due to a lower quality of our encoder $\phi$ and improving it offers an interesting direction of further research. While both the encoder and decoder together within an AE-ROM may produce a good reconstruction, it may be beneficial to ensure some  balance of the learned dynamics of the two halves of the AE-ROM system. This could be achieved by enforcing that the  nonlinear transformations performed in the two halves are roughly mirror images of each other, rather than potentially allowing for either the decoder or encoder to handle the bulk of the  reconstruction process. This may allow for better behavior of both the encoder and decoder when used separately, rather than producing good results only when used in tandem.

\section*{Acknowledgements}
This work was supported 
by DOE through grant ASCR DE-SC0021313, and by the Computational Science Laboratory at Virginia Tech.

\frenchspacing
\fontsize{9.0pt}{10.0pt}\selectfont
\bibliography{biblio}
\end{document}